\documentclass[11pt]{article}

\usepackage[final]{acl}

\usepackage{times}
\usepackage{latexsym}

\usepackage[T1]{fontenc}

\usepackage[utf8]{inputenc}

\usepackage{microtype}

\usepackage{inconsolata}

\usepackage{graphicx}

\usepackage{amsmath}
\usepackage{amssymb}
\usepackage{mathtools}
\usepackage{amsthm}
\usepackage{array}
\usepackage{siunitx}
\usepackage{multirow}
\usepackage{graphicx}
\usepackage{subcaption}
\usepackage{booktabs}

\title{RADAR: Accelerate Large Language Model Inference with RL-Based Dynamic Draft Trees}

\author{
  \textbf{Junjie Ma}\textsuperscript{1} \quad
  \textbf{Jinlong Li}\textsuperscript{1}\thanks{Corresponding author.} \quad
  \textbf{Jiajun Luo}\textsuperscript{2} 
  \\
  \textsuperscript{1}University of Science and Technology of China \\
  \textsuperscript{2}Tsinghua University \\
  \texttt{mahiru@mail.ustc.edu.cn} \quad
  \texttt{jlli@ustc.edu.cn} \quad
  \texttt{luo-jj24@mails.tsinghua.edu.cn}
}

\begin{document}
\maketitle
\begin{abstract}
Inference with modern Large Language Models (LLMs) is expensive and slow, and speculative sampling has emerged as an effective solution to this problem. However, the number of calls to the draft model for generating candidate tokens in speculative sampling is a preset hyperparameter, lacking flexibility. To generate and utilize the candidate tokens more effectively, we propose RADAR, a novel speculative sampling method with RL-based dynamic draft trees. RADAR formulates the draft tree generation process as a Markov Decision Process (MDP) and employs offline reinforcement learning to train a prediction model, which enables real-time decisions on calls to the draft model, reducing redundant computations and further accelerating inference. Evaluations across three LLMs and four tasks show that RADAR achieves a speedup of $3.17$x--$4.82$x over the auto-regressive decoding baseline. The code is available at \url{https://github.com/minaduki-sora/RADAR}.
\end{abstract}

\section{Introduction}

Modern Large Language Models (LLMs) exhibit remarkable capabilities and are widely applied across various tasks~\cite{achiam2023gpt,guo2025deepseek}. The parameter sizes of LLMs range from tens of billions to even hundreds of billions, and during auto-regressive generation, each generated token requires accessing entire model parameters, making LLM inference slow and costly. Here we focus on accelerating LLM inference.

To accelerate LLM inference, some methods have been proposed recently, such as prompt pruning~\cite{zhou2023efficient}, quantization~\cite{frantar2022gptq}, knowledge distillation~\cite{agarwal2024policy}, speculative sampling~\cite{leviathan2023fast,chen2023accelerating} and so on. Therein, as one of the most popular methods, speculative sampling uses a small draft model to quickly generate candidate tokens, which are then verified by target LLM in parallel. Speculative sampling allows multiple tokens to be generated in a single LLM pass, significantly reducing inference latency while preserving the original output distribution of target LLM. For speculative sampling, the recent studies~\cite{miao2024specinfer, li2024eagle, gao2025falcon, li2024eagle2, zhang2024learning} mostly focus on organizing and utilizing the tokens generated by the draft model more effectively.

In speculative sampling, the generated tokens are generally organized in a certain data structure, termed as \textit{draft structure} here. And there are two types of draft structures: chain and tree-based. Speculative sampling~\cite{leviathan2023fast,chen2023accelerating} first employs an auto-regressive method to invoke the draft model and produce a sequence of candidate tokens, which is referred to as a chain-structured draft. To promote parallelism during LLM’s verifications, candidate tokens are organized as a verification tree~\cite{miao2024specinfer}, which is then serialized and verified via masked attention in a single LLM pass.
However, most draft trees ~\cite{miao2024specinfer,li2024eagle,cai2024medusa,ankner2024hydra} are static during token generation across all contexts or tasks. 
By reordering and pruning nodes of draft tree, methods like EAGLE-2~\cite{li2024eagle2} and EAGLE-3~\cite{li2025eagle3} 
attempt to improve the acceptance length. However, 
the number of calls to the draft model remains a hyperparameter, lacking flexibility. 
For instance, we run the source code provided by EAGLE-3 on MT-bench~\cite{zheng2023judging} and LLaMA-Instruct 3.1 8B~\cite{dubey2024llama}, results show that draft tokens are completely rejected at a frequency of about $31\%$, yet the draft model is still called 8 times~\cite{li2025eagle3}. While methods like SpecDec++~\cite{huang2024specdec++} and DISCO~\cite{mamou2024accelerating} attempt to dynamically adjust the draft length by predicting the acceptance probability for each node, their methodologies are strictly designed for chain-based drafts and cannot be generalized to recent state-of-the-art tree-based methods such as EAGLE-3.
Here, we attempted to dynamically decide the calls to the draft model during the draft stage, thereby generating draft trees with varying depths based on different contexts.

We propose to decide whether to invoke the draft model at each step of draft generation process, rather than predicting the required depth of the draft tree.
This is because the depth of the draft tree, equivalent to the acceptance length, is stochastic due to the rejection sampling in the speculative sampling~\cite{leviathan2023fast,chen2023accelerating}, making it impossible to obtain labeled data. However, we observed that the draft tree generation process can be modeled as a Markov Decision Process (MDP), which eliminates the need for labeled data by utilizing intrinsic and extrinsic rewards. We propose to calculate these rewards based on acceptance length distributions of draft trees, 
which are obtained by running speculative sampling algorithm on shareGPT\footnote{https://huggingface.co/datasets/anon8231489123/ShareGPT\\\_Vicuna\_unfiltered\label{shareGPT}} dataset, and then all these distributions are collected to construct a dataset for training the prediction model. 
During training, the prediction model samples policy trajectories in real time based on the real MDP, which enables offline reinforcement learning without extrapolation error~\cite{fujimoto2019off}.

Upon these insights, we propose a Reinforcement learning Adjusted Draft-generation AlgoRithm for speculative sampling (RADAR), leveraging the confidence scores from the draft model to dynamically decide the calls 
to the draft model. In summary, our key contributions include:

\begin{itemize}
\item We introduce a speculative sampling framework with dynamic draft trees, namely RADAR, which employs a lightweight prediction model to adaptively decide the  calls to the draft model during the draft stage.

\item 
To handle the lack of labeled data for training the prediction model, we model the draft tree generation process as a Markov Decision Process (MDP), and address it through a newly introduced dataset of acceptance length distribution, from which an acceptance length is sampled and then used for the calculation of action reward, 
eliminating the need for expensive real-time interactions with the LLM.
\end{itemize}

\section{Related Work}
\subsection{Speculative decoding}
Since the proposal of speculative decoding, researchers have been improving the algorithm from different perspectives. One line of research focuses on enhancing alignment between draft and target models via distillation~\cite{1019494} or feature-level integration~\cite{li2024eagle}. Another direction involves optimizing draft structures, shifting from chain-based to tree-based decoding~\cite{miao2024specinfer} and utilizing pruning and reranking~\cite{li2024eagle2} to increase the acceptance rate. The third targets system parallelism to minimize drafting latency, such as Medusa’s~\cite{cai2024medusa} parallel decoding heads and PEARL’s~\cite{liu2025pearl} framework that overlaps the drafting and verification phases asynchronously. Our work advances the latter by introducing a dynamic draft tree that adaptively adjusts the number of calls to the draft model, resulting in minimizing drafting latency. Crucially, our approach is orthogonal to PEARL: while PEARL achieves dynamic candidate length, its primary contribution and motivation lie in maximizing temporal parallelism between draft and verification via asynchronous execution. In contrast, RADAR focuses on the redundant calls to the draft model and make real-time decisions to adjust the draft tree. These two paradigms can be integrated to achieve even better acceleration.

\subsection{Dynamic draft structure}
A major efficiency bottleneck in conventional speculative decoding methods stems from their inherent dependency on a fixed hyperparameter K, representing the number of calls to the draft model. Vanilla Speculative Sampling~\cite{leviathan2023fast} derives an optimal $K$ based on the i.i.d. assumption of token acceptance probabilities. To introduce context-awareness, Kangaroo~\cite{liu2024kangaroo} adopts a simple heuristic that terminates speculation if the current draft token's confidence falls below a threshold. Going a step further, SpecDec++~\cite{huang2024specdec++} employs a trained prediction head to estimate individual acceptance probabilities and stops speculation based on their calculated cumulative product. In a similar vein, the primary distinction in DISCO~\cite{mamou2024accelerating} is the use of a specialized head to directly predict this cumulative product, bypassing the need for manual multiplication of individual probabilities. Despite these advancements, previous methods~\cite{liu2024kangaroo, huang2024specdec++, mamou2024accelerating} are inherently designed for chain-based draft, making them difficult to generalize to more complex tree-based draft structures that involve pruning and reranking~\cite{li2024eagle2}. Furthermore, these approaches typically rely on supervised learning to estimate acceptance probabilities and use a predefined threshold to implement termination, which remains a rigid hyperparameter. In contrast, RADAR addresses these limitations by modeling the speedup ratio end-to-end using reinforcement learning. This framework enables context-aware dynamic drafting that is compatible with state-of-the-art tree-based speculative decoding methods, effectively accelerating inference across diverse contexts.

Concurrent with our work, LTD~\cite{zhang2026learning} also addresses the efficiency issue caused by redundant draft calls in speculative decoding, and likewise applies reinforcement learning to dynamic draft tree control. Different from RADAR, LTD relies on online reinforcement learning and requires costly interaction with the target LLM during training, whereas RADAR adopts an offline reinforcement learning framework that avoids such online interaction. We include an empirical comparison with LTD in the main experiments, and further discuss differences in Appendix~\ref{appdendix:ltd}.

\subsection{EAGLE Series}
Distinct from vanilla speculative decoding that employs a separate small language model for token-level drafting, EAGLE~\cite{li2024eagle} introduces auto-regression at the feature level. It predicts the second-to-top layer hidden states of the target model $M_p$ and reuses $M_p$'s original LM head for sampling, which mitigates the distribution shift between the draft and target models. 

The framework has evolved through two major iterations. EAGLE-2~\cite{li2024eagle2} identifies that acceptance rates are context-dependent rather than purely position-dependent. It leverages the draft model's confidence scores as a proxy for acceptance probability to construct a context-aware draft tree, reordering and pruning nodes at runtime to optimize the verification efficiency. More recently, EAGLE-3~\cite{li2025eagle3} addresses the limitations of feature prediction and scaling. By shifting from pure feature extrapolation to direct token prediction and incorporating \textit{multi-layer feature fusion}, EAGLE-3 significantly enhances drafting accuracy and benefits from large-scale data training. 

Despite these advancements, the drafting process in the EAGLE family remains constrained by a fixed number of draft model calls. As observed in our preliminary experiments, this static hyperparameter often leads to redundant drafting efforts, highlighting a lack of flexibility in adapting to real-time verification.

\section{Preliminaries}
\textbf{Speculative Sampling}~\cite{leviathan2023fast, chen2023accelerating} has emerged as a standard technique to accelerate the inference of LLMs without compromising output quality. The core mechanism involves a collaboration between a computationally efficient draft model $M_q$ and a powerful target model $M_p$. Given an input prefix $\mathbf{x}$, the process begins with a drafting phase, where $M_q$ auto-regressively predicts $t$ future tokens. Specifically, at each drafting step $i \in \{1, \dots, t\}$, a candidate token $\hat{x}_i$ is sampled from the distribution $q_i(\cdot) = M_q(\cdot \mid \mathbf{x}, \hat{x}_{1}, \dots, \hat{x}_{i-1})$. This stage results in a draft sequence $\hat{\mathbf{X}} = \{\hat{x}_1, \dots, \hat{x}_t\}$ along with its corresponding draft probabilities.

Following the drafting phase, the target model $M_p$ performs a single parallel forward pass on the concatenated sequence $[\mathbf{x}, \hat{x}_1, \dots, \hat{x}_t]$ to compute the target distributions $p_1, \dots, p_{t+1}$, where $p_i = M_p(\cdot \mid \mathbf{x}, \hat{x}_1, \dots, \hat{x}_{i-1})$. To ensure the output follows the target distribution $p$ exactly, a speculative sampling criterion is applied sequentially. For each candidate $\hat{x}_i$, it is accepted with a probability as Eq.~(\ref{prob}):
\begin{equation}\label{prob}
    \alpha_i = \min\left(1, \frac{p_i(\hat{x}_i)}{q_i(\hat{x}_i)}\right).
\end{equation}
If the token is accepted, the verification proceeds to $\hat{x}_{i+1}$; otherwise, the process is truncated at index $i$, and a new token is resampled from the rectified distribution as Eq.~(\ref{recited}):
\begin{equation}\label{recited}
    p'_i(x) = \frac{\max(0, p_i(x) - q_i(x))}{\sum_{x'} \max(0, p_i(x') - q_i(x'))}.
\end{equation}
If all $t$ tokens are accepted, an additional token is sampled from $p_{t+1}$ to maximize efficiency. This ensures that in each iteration, the system generates between $1$ and $t+1$ tokens, while the final output remains statistically identical to the distribution produced by the target model alone, thereby maintaining the quality of the generation.

\begin{figure}[htb]

\begin{minipage}[b]{1.0\linewidth}
  \centering
  \centerline{\includegraphics[width=\columnwidth]{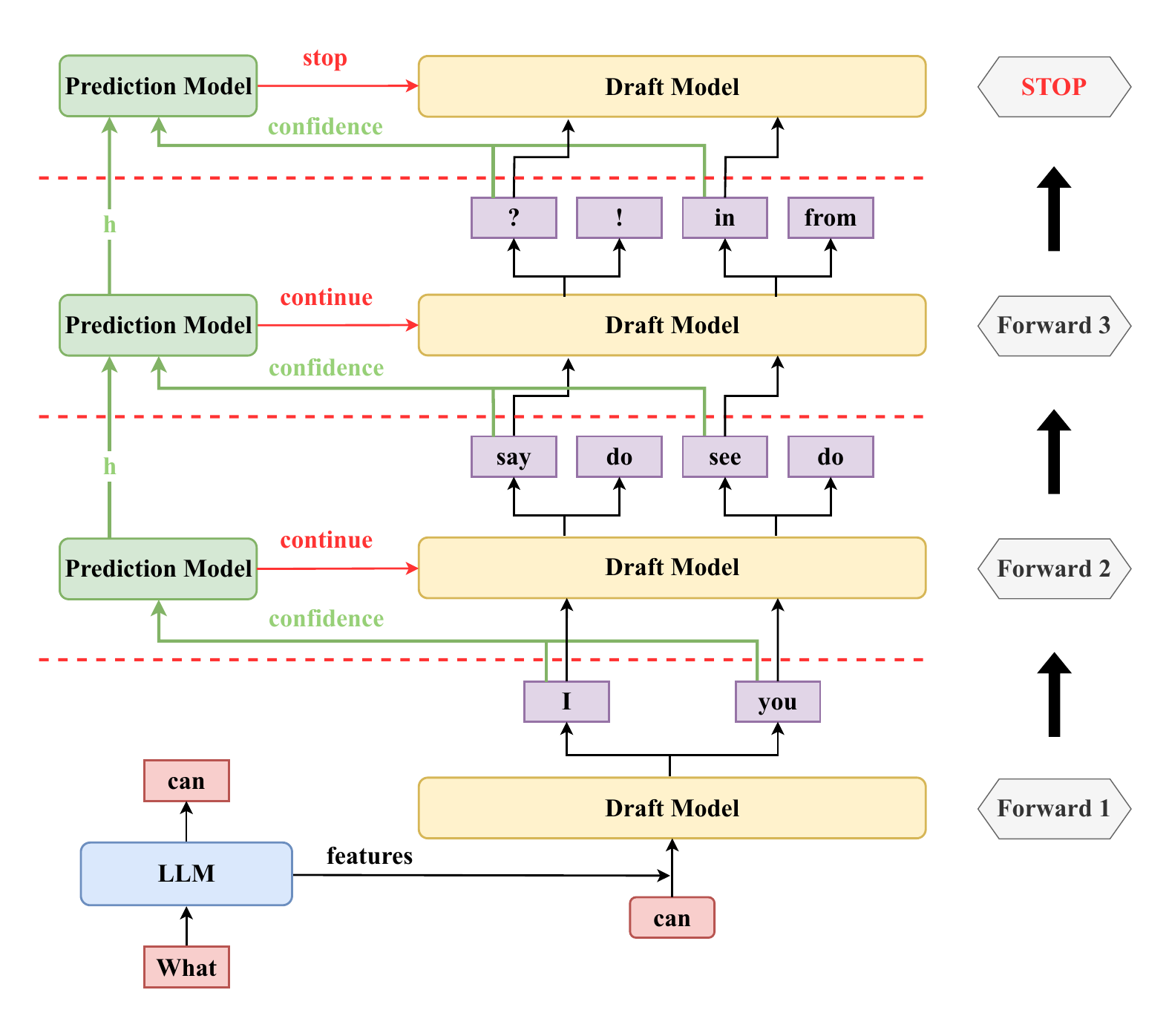}}
  \centerline{(a) RADAR}\medskip
\end{minipage}
\begin{minipage}[b]{1.0\linewidth}
  \centering
  \centerline{\includegraphics[width=\columnwidth]{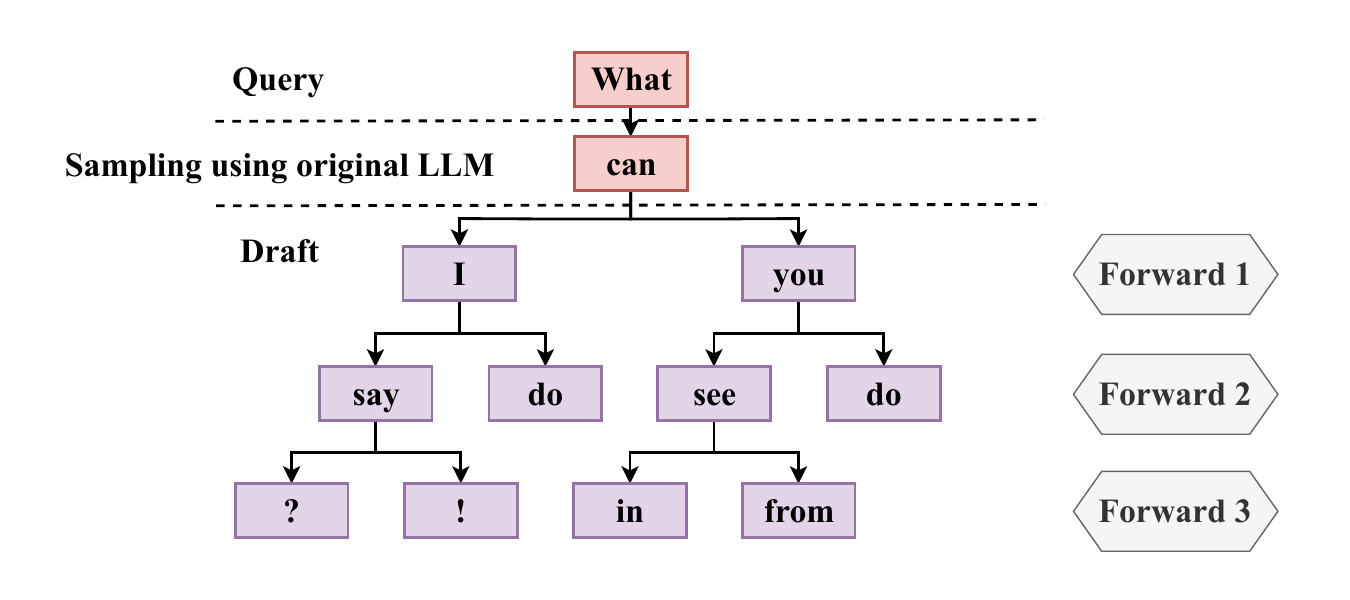}}
  \centerline{(b) Draft Tree}\medskip
\end{minipage}
\caption{Pipeline of RADAR. Figure (a) illustrates the steps of the prediction model controlling the draft model generation, while the Figure (b) demonstrates the corresponding generation results.}
\label{fig:radar}
\end{figure}

\section{RADAR}

RADAR consists of three components: the target LLM, the draft model, and the prediction model, as illustrated in Figure \ref{fig:radar} (a).
Following the description of the inference pipeline, we explain the \textit{prediction model}, the core component, in detail.

\subsection{Inference Pipeline}
Compared with previous popular speculative sampling methods such as EAGLE-3~\cite{li2025eagle3}, we introduce a prediction model that dynamically decides to \textit{continue/stop} the draft model's auto-regressive generation at each generation step. Here, we illustrate the inference pipeline of our approach by example, as shown in Figure \ref{fig:radar}. A detailed step-by-step walkthrough of this inference pipeline is provided in Appendix~\ref{appendix:pipeline}.


\subsection{Prediction Model}

The primary goal of the prediction model is to dynamically decide the calls to the draft model during the draft phase, enabling the construction of a dynamic draft tree. Considering the sequential nature of the generation process, our prediction model is designed as an LSTM-based network, which takes a hidden vector initialized to zero and the top-$k$ confidence scores from the draft model as inputs. While the MLP policy uses only the confidence scores as its state, the LSTM policy additionally maintains a recurrent hidden state that summarizes previous drafting steps. We use the confidence scores as inputs because there is a strong positive correlation between the draft model's confidence score and the acceptance rate of the token~\cite{li2024eagle2}. The model then outputs a binary control signal \textit{continue/stop}, recorded as \textit{logits}, and an updated hidden vector $h$.

We elaborate our prediction model on two aspects: problem modeling and model training.


\textbf{Problem modeling.} We formulate the decision of calls to the draft model as a Markov Decision Process (MDP), defined by a tuple $\mathcal{M}=(S,A,P,r,\gamma)$. The detailed definitions of these elements are as Eqs. (\ref{eq:state}-\ref{eq:gamma}):

\begin{itemize}
    \item \textbf{State Space ($S$):} The state $s_t \in S$ is defined as the confidence scores of the top-$k$ candidate tokens generated by the draft model at time step $t$, as Eq.~(\ref{eq:state}):
    \begin{equation}
        s_t = [c_{1}, c_{2}, \dots, c_k], \quad 1\le t \le t_{\text{max}}, \label{eq:state}
    \end{equation}
    where $c_i \in [0,1]$, $k$ denotes the branching factor of the draft tree, and $t_{\text{max}}$ is the maximum number of calls to the draft model.

    \item \textbf{Action Space ($A$):} The action space consists of two actions, as defined in Eq.~(\ref{eq:a}):
    \begin{equation}\label{eq:a}
        A=\{0,1\}.
    \end{equation}
    Action "$0$" indicates the termination of draft generation, while action "$1$" indicates the continuation to generate candidate tokens.

    \item \textbf{Transition Probability ($P$):} The state transition function $P(s_{t+1} | s_t, a_t)$ is defined as Eq.~(\ref{eq:tran}):
    \begin{equation}
        P(s_{t+1} | s_t, a_t) =
        \begin{cases}
        \delta(s_{t+1} = s_{\bot}) & \text{if } a_t = 0 \\
        P_{\text{draft}}(s_{t+1} | s_t) & \text{if } a_t = 1
        \end{cases}, \label{eq:tran}
    \end{equation}
    where $s_{\bot}$ is an absorbing terminal state reached when $a_t = 0$, and $P_{\text{draft}}(\cdot)$ represents the transition probability distribution based on the draft model's generation when $a_t = 1$.

    \item \textbf{Reward Function ($r$):} The reward $r_t$ imposes a constant penalty $-\alpha$ at each step to discourage excessive drafting, and provides a final reward estimating the inference speed upon termination. A detailed interpretation of $\alpha$ is provided in Appendix~\ref{appendix:alpha}. The reward at time step $t$ is defined as Eq.~(\ref{eq:reward}):
    \begin{equation}
        r_t =
        \begin{cases}
        -\alpha & \text{if } t < T \\
        \dfrac{\ell_{\text{acc}}}{T_{\text{gen}}(T)} & \text{if } t = T
        \end{cases}, \label{eq:reward}
    \end{equation}
    where $\ell_{\text{acc}}$ is the acceptance length. $T_{\text{gen}}(t)$ estimates the total time cost, defined as Eq.~(\ref{eq:tgen}):
    \begin{equation}\label{eq:tgen}
        T_{\text{gen}}(t) = T_{\text{o}} + T_{\text{d}} \cdot t + T_{\text{pred}}(t),
    \end{equation}
    \begin{equation}\label{eq:tdyn}
        T_{\text{pred}}(t) =
        \begin{cases}
            T_{\text{p}} \cdot (t + 1) & \text{if } t < t_{\text{max}} \\
            T_{\text{p}} \cdot t & \text{if } t = t_{\text{max}}
        \end{cases},
    \end{equation}
    where $T_{\text{o}}$ denotes a fixed overhead, $T_{\text{d}}$ denotes the latency of a single forward pass of the draft model, $T_{\text{p}}$ denotes the latency of a single forward pass of the prediction model, and $T_{\text{pred}}$ is the inference cost of the prediction model, which is different at the maximum step $t_{\text{max}}$, as the draft tree generation terminates, thus eliminating the need for a subsequent call to the prediction model. A breakdown of latency is provided in Appendix~\ref{appendix:latency}.

    \item \textbf{Discount Factor ($\gamma$):} We set the discount factor to balance immediate and future rewards, following standard reinforcement learning practices~\cite{mnih2015human}, as Eq.~(\ref{eq:gamma}):
    \begin{equation}
        \gamma=0.99. \label{eq:gamma}
    \end{equation}
\end{itemize}

\textbf{Model training.}
The training loss is the negative expected reward, as Eq. (\ref{eq:loss}):  
\begin{equation}
    \mathcal{L}(\theta) = -E_{\tau \sim \pi_{\theta}} [\sum_{t=0}^{T-1} (\sum_{t\prime =t}^{T} \gamma^{t\prime - t} r_{t\prime}) \log \pi_{\theta}(a_t|s_t)]
    \label{eq:loss},
\end{equation}
where $\tau = (s_1,a_1,...,s_T)$ is a trajectory sampled from policy $\pi_\theta$.

\subsection{Offline Dataset}

We construct an offline dataset to train the prediction model to avoid expensive real-time interactions with the LLM during training.
Following the description of the dataset and its construction, we explain the use of our dataset.

For each data point $x$ in the dataset, $x=\{\mathcal{S}, \mathcal{D} \}$, where $\mathcal{S}=\{s_1,s_2,\dots, s_{t_{\text{max}}}\}$ denotes the sequence of states, where $s_t$ and $t_{\text{max}}$ are defined by Eq. (\ref{eq:state}); $\mathcal{D}=\{d_1,d_2,\dots,d_{t_{\text{max}}} \}$ denotes the set of distributions for acceptance lengths corresponding to different numbers of calls to the draft model. Specifically, for each $i$, $d_i = \{p_0, p_1, \dots, p_{t_{\text{max}}}\} $ where $p_j$ denotes the probability that the acceptance length is j given that i calls were made to the draft model.

To construct the dataset, we run EAGLE-3 on ShareGPT
\footref{shareGPT}
dataset, and for each prefix in the corpus, we repeat the following procedure.
First, during the draft stage, we enumerate the number of calls to the draft model from $1$ to $t_{\text{max}}$, to generate multiple draft trees. The confidence scores of the top-$k$ tokens at each generation step are collected to form the state sequence $\mathcal{S}$. Next, draft trees are serialized and verified in parallel by the LLM, and during the verification phase, for each $i$, the distribution $d_i$ of acceptance lengths for the draft tree generated with $i$ calls is derived. These distributions are then aggregated into the set $\mathcal{D}$ and we can obtain one data point $x=\{\mathcal{S}, \mathcal{D} \}$.

\begin{figure}[ht]
  \vskip 0.1in
  \begin{center}
    \centerline{\includegraphics[width=0.5\columnwidth]{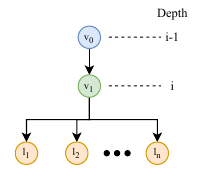}}
    \caption{
      Demonstration of the node in a draft tree for the derivation of the acceptance length distribution. The node $v_1$ has parent $v_0$ and children $l_1, l_2, \dots, l_n$, with a depth of $i$.
    }
    \label{fig:node}
  \end{center}
\end{figure}

Here, we detail the derivation of the acceptance length distribution, defined by $d_i = \{p_0, p_1, \dots, p_{t_{\text{max}}}\} $.
For a draft tree node $v_1$ with parent $v_0$ and children $l_1, l_2, \dots, l_n$, as shown in Figure \ref{fig:node}, the probability $p_i$ used in the dataset is defined by Eqs. (\ref{eq:dist}-\ref{eq:prej}). 
\begin{align}
    &p_i = \sum_{v.\text{depth} = i} P_{\text{stop}}(v)\label{eq:dist},
    \\
    &P_{\text{stop}}(v_1) = P_{\text{acc}}(v_0) \cdot P_{\text{acc}}(v_1 | v_0) \cdot P_{rej}(l|v_1) \label{eq:pstop},
    \\
    &P_{rej}(l|v_1) = 1 - \sum_{j=1}^n P_{\text{acc}}(l_j | v_1) \label{eq:prej},
\end{align}
where $P_{\text{stop}}(v_1)$ denotes the probability that $v_1$ is accepted while all of its children are rejected by LLM during verification stage; $P_{\text{acc}}(v_0)$ denotes the probability that node $v_0$ is accepted, which can be calculated based on speculative sampling algorithm~\cite{leviathan2023fast,chen2023accelerating}; \(P_{rej}(l|v_1)\) denotes the probability that all children are rejected given that their parent $v_1$ is accepted; $P_{\text{acc}}(v_1 | v_0)$ denotes the probability that the child node $v_1$ is accepted given that its parent $v_0$ has been accepted, and $p_i$ denotes the probability that the acceptance length equals $i$.

To use the dataset for training prediction model, for each data point $x=\{\mathcal{S}, \mathcal{D} \}$, the state sequence $\mathcal{S}$ is input into the prediction model, which produces a complete trajectory denoted as $\tau = (s_1, a_1, \dots, s_T)$. From this trajectory, the number of calls to the draft model, denoted by $i$, is obtained. Subsequently, the acceptance length $\ell_{\text{acc}}$ is sampled from the distribution $d_i \in \mathcal{D}$. The reward and loss are computed using the reward function defined in Eq. (\ref{eq:reward}) and the loss function defined in Eq. (\ref{eq:loss}), respectively. Finally, the parameters of the prediction model are updated using policy gradient methods.
Through the above steps, both the data distribution and the behavior of the predictive model in offline reinforcement learning remain consistent with those in online reinforcement learning, thereby eliminating the impact of extrapolation error on the training process~\cite{fujimoto2019off}.

\begin{table*}[t]
\centering
\begin{tabular}{ll *{4}{S[table-format=1.2x] S[table-format=1.2]}}
\toprule
\textbf{Model} & \textbf{Method} & 
\multicolumn{2}{c}{\textbf{MT-bench}} & 
\multicolumn{2}{c}{\textbf{GSM8K}} & 
\multicolumn{2}{c}{\textbf{Alpaca}} & 
\multicolumn{2}{c}{\textbf{MBPP}} \\
\cmidrule(lr){3-4} \cmidrule(lr){5-6} \cmidrule(lr){7-8} \cmidrule(lr){9-10}
 & & {Speedup} & {$\tau$} & {Speedup} & {$\tau$} & {Speedup} & {$\tau$} & {Speedup} & {$\tau$} \\
\midrule
\multirow{6}{*}{L3 8B} 
    & Eagle-2 & 2.56x & 3.37 & 3.43x & 3.88 & 2.89x & 4.08 & 3.29x & 4.55 \\
    & Eagle-3 & 3.08x & 4.55 & 4.68x & 5.50 & 3.86x & 5.62 & 4.21x & 6.12 \\
    & Eagle-3+DISCO & 3.10x & 4.07 & 4.51x  & 5.08 & 3.68x & 5.32 & 3.71x & 5.74 \\
    & Eagle-3+Spec++ & 3.12x & 4.40 & 4.44x  & 5.39 & 3.63x & 5.36 & 3.98x & 5.81\\
    & Eagle-3+LTD & 2.98x & 5.12 & 4.61x & 6.04 & 3.72x & 6.47 & 4.28x & 6.89 \\
    & Eagle-3+RADAR & \textbf{3.41x} & 4.48 & \textbf{4.82x} & 5.32 & \textbf{4.04x} & 5.51 & \textbf{4.44x} & 6.00 \\
\addlinespace
\multirow{6}{*}{V 13B}
    & Eagle-2 & 2.89x & 4.38 & 3.18x & 4.46 & 2.83x & 4.09 & 3.56x & 4.93 \\
    & Eagle-3 & 3.74x & 5.69 & 4.24x & 5.94 & 3.50x & 5.51 & 4.55x & 6.44 \\
    & Eagle-3+DISCO & 3.68x & 5.67 & 3.64x  & 5.86 & 3.47x & 5.63  & 4.25x & 6.62\\
    & Eagle-3+Spec++ & 3.60x & 5.54 & 3.64x  & 5.81 &  3.37x & 5.61 & 4.38x & 6.45 \\
    & Eagle-3+LTD & 3.78x & 6.62 & 3.82x & 6.60 & 3.31x & 6.12 & 4.56x & 7.98 \\
    & Eagle-3+RADAR & \textbf{4.05x} & 5.67 & \textbf{4.36x} & 5.87 & \textbf{3.84x} & 5.48 & \textbf{4.75x} & 6.42 \\
\addlinespace
\multirow{3}{*}{DSL 8B}
    & Eagle-3 & 3.42x & 4.88 & 4.39x & 6.39 & 3.08x & 4.49 & 3.71x & 5.35 \\
    & Eagle-3+LTD & 3.56x & 4.16 & 4.15x & 4.61 & \textbf{3.39x} & 4.03 & 3.97x & 4.40 \\
    & Eagle-3+RADAR & \textbf{3.86x} & 4.85 & \textbf{4.71x} & 6.33 & 3.17x & 4.44 & \textbf{3.99x} & 5.31 \\
\bottomrule
\end{tabular}
\caption{Speedup ratios and average acceptance lengths $\tau$ of different methods setting temperature=1.
L3 represents LLaMA-Instruct 3.1, V represents Vicuna and DSL represents DeepSeek-R1-Distill-LLaMA. Spec++ stands for SpecDec++.
The bold text in the speedup ratio column indicates the highest value for each dataset and model.
}
\label{tab:model_metrics}
\end{table*}

\section{Experiment}

\subsection{Setup}
\textbf{Target LLMs.} We conduct experiments with state-of-the-art open-source chat and reasoning models, including LLaMA-Instruct 3.1 8B~\cite{dubey2024llama}, Vicuna 13B~\cite{chiang2023vicuna} and DeepSeek-R1-Distill-LLaMA 8B~\cite{guo2025deepseek}. 

\textbf{Tasks.} Following EAGLE~\cite{li2024eagle} and HASS~\cite{zhang2024learning}, we conduct evaluations on four common tasks, using the same weights for all tasks without fine-tuning on the respective tasks. For multi-turn conversation, mathematical reasoning, instruction following and code generation, we chose the MT-bench~\cite{zheng2023judging}, GSM8K~\cite{cobbe2021training}, Alpaca~\cite{taori2023stanford} and MBPP~\cite{austin2021program} datasets, respectively.

\textbf{Metrics}. RADAR does not fine-tune the target LLM's weights and adopts a strict speculative sampling algorithm, ensuring no loss in performance. Thus, we do not evaluate generation quality. We use the following two metrics to assess acceleration performance:
\begin{itemize}
    \item \textbf{Speedup Ratio:} The actual test speedup ratio relative to vanilla auto-regressive decoding.

    \item \textbf{Average Acceptance Length $\tau$:} The average number of tokens generated per drafting-verification cycle, which corresponds to the number of tokens accepted from the draft.

\end{itemize}


\textbf{Comparisons.} We use vanilla auto-regressive decoding as the baseline, serving as the benchmark for speedup ratios (1.00x). We compare RADAR with recent lossless speculative decoding methods, including EAGLE-2~\cite{li2024eagle2} and EAGLE-3~\cite{li2025eagle3}, as well as recent adaptive drafting methods, including SpecDec++~\cite{huang2024specdec++}, DISCO~\cite{mamou2024accelerating}, and LTD~\cite{zhang2026learning}.

\textbf{Implementation.} Our code is built based on EAGLE-3's open-source repository. During training, we adopt the REINFORCE algorithm~\cite{williams1992simple} for training prediction model, and set the learning rate to 1e-4. During evaluation, following EAGLE-3's setup, $k$ is set to 10, the maximum number of calls to the draft model is 8, temperature is set to 1.0 and batch size is set to 1.  More details are provided in Appendix~\ref{appendix:implementation}.

\subsection{Effectiveness of RADAR}
Note that the speedup ratio is hardware-sensitive due to variations in computational capabilities, most of the previous studies~\cite{leviathan2023fast,chen2023accelerating,miao2024specinfer,li2024eagle} performed inference tests in the same environment. Thereby, in our experiments,  all inference tests were run on 2$\times$ NVIDIA RTX3090 GPUs. We additionally report results on LLaMA-Instruct 3.3 70B using 2$\times$RTX PRO 6000 GPUs in Appendix~\ref{appendix:70b}.

As shown in Table \ref{tab:model_metrics}, we present different methods' speedup ratios and acceptance lengths across four datasets. RADAR provides a speedup of approximately $3.17$x--$4.82$x compared to vanilla auto-regressive generation, with a $3\%$--$12.9\%$ improvement over EAGLE-3. Besides, RADAR maintains a high average acceptance length, only about $1.2\%$ lower than that of EAGLE-3. On most of the tested tasks and target LLMs, RADAR achieves the highest speedup ratio.

Generally, given the same model and dataset, a higher acceptance length implies a greater speedup ratio, however, RADAR deviates from this pattern by reducing redundant draft calls. 
As shown in Table \ref{tab:lambda_counts}, we present the average number of calls to the draft model per draft-verification cycle across four different datasets with three LLMs as target model, respectively. Under the configuration where the maximum number of calls is set to 8, RADAR reduces the average number of calls by $9.3\%$--$34.3\%$ compared to EAGLE-3’s fixed 8 calls, with an average reduction of $18.7\%$. RADAR effectively decreases the frequency of draft model invocations and reduces the time overhead during the draft phase.

\begin{table}[ht]
\centering
\small 
\begin{tabular}{l *{4}{S[table-format=2.2]}}
\toprule
\textbf{Model} & \textbf{MT-bench} & \textbf{GSM8K} & \textbf{Alpaca} & \textbf{MBPP} \\
\midrule
L3 8B  & 5.25 & 6.19 & 6.20 & 6.60 \\
V 13B  & 6.88 & 7.26 & 6.83 & 7.26 \\
DSL 8B & 6.10 & 7.20 & 5.85 & 6.47 \\
\bottomrule
\end{tabular}
\caption{Average number of calls to draft model of RADAR under temperature=1.0. L3 represents LLaMA-Instruct 3.1, V represents Vicuna and DSL represents DeepSeek-R1-Distill-LLaMA. Following the original papers' setup, EAGLE-3 has an preset number of calls of 8 across all tested models and datasets. }
\label{tab:lambda_counts}
\end{table}

\subsection{Case Study of Acceptance Length and Draft-Model Calls}

\begin{figure}[t]
\vskip 0.1in
\begin{minipage}[b]{1.0\linewidth}
  \centering
  \centerline{\includegraphics[width=\columnwidth]{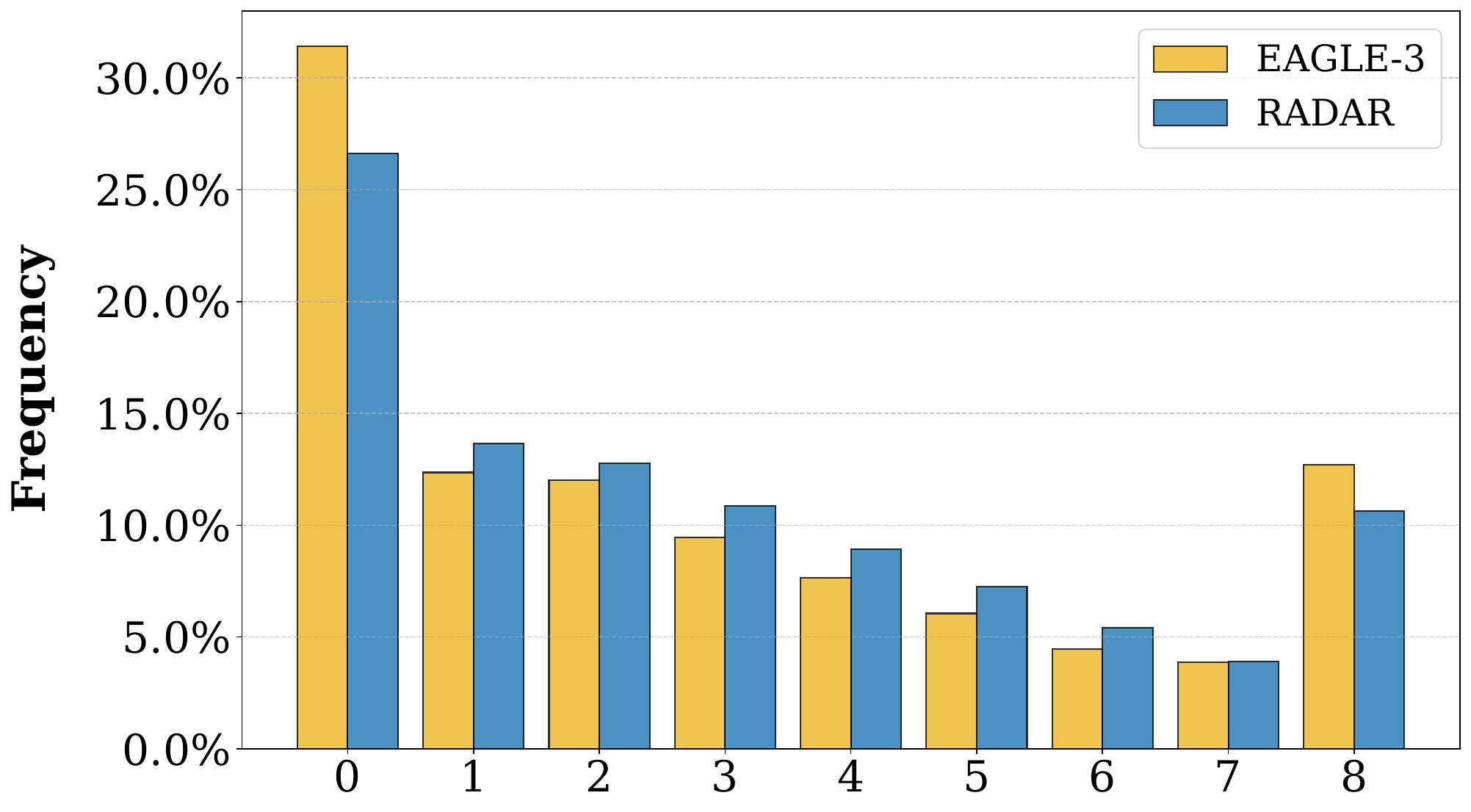}}
  \centerline{(a) The acceptance length of RADAR and EAGLE-3.}\medskip
\end{minipage}
\begin{minipage}[b]{1.0\linewidth}
  \centering
  \centerline{\includegraphics[width=\columnwidth]{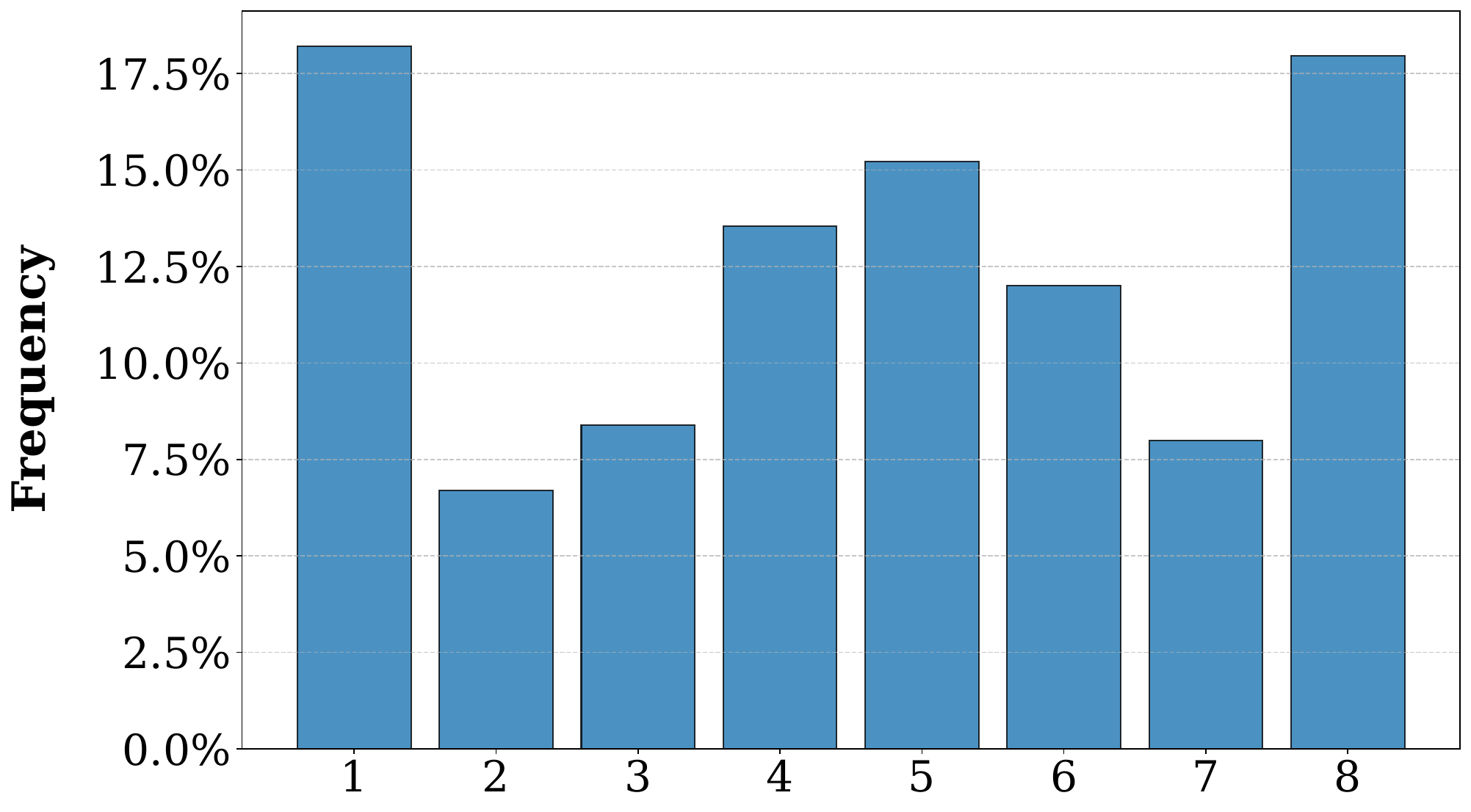}}
  \centerline{(b) The number of calls to the draft model of RADAR.}\medskip
\end{minipage}
\caption{Distribution of acceptance length and draft-model calls on MT-bench with LLaMA-Instruct 3.1 8B as the target model under temperature $=1.0$.}
\label{fig:ac}
\end{figure}

Figure~\ref{fig:ac} illustrates the distribution of acceptance length and the number of calls to the draft model on MT-bench, using LLaMA-Instruct 3.1 8B as the target model. As shown in Figure~\ref{fig:ac}(a), while maintaining a high acceptance length, RADAR significantly reduces the probability of an acceptance length of 0 compared with EAGLE-3. It also increases the frequency of acceptance lengths from 1 to 7, while the frequency at length 8 decreases slightly. This is because the dynamic drafting mechanism not only reduces draft-model calls but also transforms the draft tree from a tall and thin structure into a shorter and wider one, thereby increasing the acceptance probability at lower depths.

Figure~\ref{fig:ac}(b) shows how RADAR dynamically determines the number of draft-model calls. The prediction model stops draft-tree generation early when further drafting is deemed inefficient. This adaptive stopping reduces redundant computation while maintaining the acceptance length, thereby accelerating inference.


\subsection{Ablation Study}

\begin{table*}[t]
\centering
\begin{tabular}{ll *{4}{S[table-format=1.2x] S[table-format=1.2]}}
\toprule
\textbf{Policy} & \textbf{Input} & 
\multicolumn{2}{c}{\textbf{MT-bench}} & 
\multicolumn{2}{c}{\textbf{GSM8K}} & 
\multicolumn{2}{c}{\textbf{Alpaca}} & 
\multicolumn{2}{c}{\textbf{MBPP}} \\
\cmidrule(lr){3-4} \cmidrule(lr){5-6} \cmidrule(lr){7-8} \cmidrule(lr){9-10}
 & & {Speedup} & {$\tau$} & {Speedup} & {$\tau$} & {Speedup} & {$\tau$} & {Speedup} & {$\tau$} \\
\midrule
MLP & scores & 3.01x & 4.55 & 4.63x & 5.26 & 3.59x & 5.30 & 4.15x & 5.87 \\
LSTM & scores & \textbf{3.41x} & 4.48 & \textbf{4.82x} & 5.32 & \textbf{4.04x} & 5.51 & \textbf{4.44x} & 6.00 \\
LSTM & hidden states & 3.10x & 4.60 & 4.74x & 5.49 & 3.89x & 5.67 & 4.29x & 6.05 \\
\bottomrule
\end{tabular}
\caption{Ablation study on different policy networks and input features evaluated on LLaMA-Instruct 3.1 8B. "Scores" indicates the confidence scores of the top-$k$ candidate tokens generated by the draft model, and "hidden states" stands for the hidden states from the draft model.}
\label{tab:ablation_inputs}
\end{table*} 

In this section, we aim to validate the effectiveness of RADAR through a series of analyses, focusing on the robustness of the reward design and the choices of policy network architectures and input features. We also provide supplementary studies on LLaMA-Instruct 3.3 70B~\cite{dubey2024llama} in Appendix~\ref{appendix:70b} for scalability, on HASS~\cite{zhang2024learning} in Appendix~\ref{appendix:hass} for generality, and across decoding temperatures in Appendix~\ref{appendix:temperature}.

\begin{figure}[ht]
  \vskip 0.2in
  \begin{center}
    \centerline{\includegraphics[width=0.9\columnwidth]{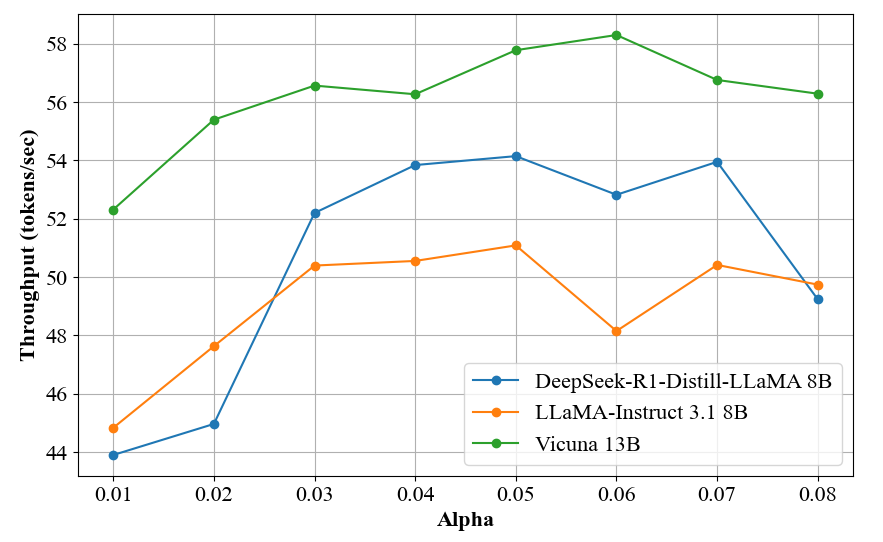}}
    \caption{
      Throughput (tokens/sec) across different $\alpha$ values on MT-bench across three LLMs. 
    }
    \label{fig:alpha_ablation}
  \end{center}
\end{figure}

\textbf{Robustness of $\alpha$.} We evaluate the impact of $\alpha$ on MT-bench across three target models, as shown in Figure~\ref{fig:alpha_ablation}. The throughput curves for all evaluated models exhibit a consistent bell-shaped trend within the range $[0.01, 0.08]$, indicating that overly small $\alpha$ values fail to sufficiently penalize redundant draft calls, while overly large values lead to premature stopping. The consistently strong performance around $\alpha \approx 0.05$ suggests that our reward design is robust across different model architectures. Furthermore, owing to the high efficiency of our offline training pipeline, the entire training process for each model finishes within 30 minutes. This efficiency makes it practical to tune $\alpha$ via a simple grid search. 

\textbf{Cross-Device Robustness.}
We evaluate the sensitivity to the device-specific latency parameter in the reward on MT-bench using LLaMA-Instruct 3.1 8B. We train three RADAR variants, denoted as RADAR-3090, RADAR-4090, and RADAR-5090, where the latency parameter is profiled on the corresponding device. As shown in Table~\ref{tab:throughput}, RADAR maintains high performance even with mismatched parameters. This is because acceptance length is hardware-independent, and the relative magnitude of $T_o$, $T_d$, and $T_p$ remains consistent across devices. While matched parameters yield the best results, device-specific tuning is trivially affordable since the training cost for the prediction model requires only about 60 TFLOPs and 30 minutes.

\begin{table}[ht]
\centering
\small
\begin{tabular}{lccc}
\toprule
\textbf{Method} & \textbf{RTX3090} & \textbf{RTX4090} & \textbf{RTX5090} \\
\midrule
Eagle-3    & 48.12          & 91.15          & 121.41         \\
RADAR-3090 & \textbf{53.87} & 99.80          & 123.02         \\
RADAR-4090 & 47.51          & \textbf{100.40} & 124.32         \\
RADAR-5090 & 51.91          & 99.10          & \textbf{131.82} \\
\bottomrule
\end{tabular}
\caption{Throughput (tokens/sec) with LLaMA-Instruct 3.1 8B as the target model across different GPUs on MT-bench. RADAR-3090, RADAR-4090 and RADAR-5090 represent models trained with latency parameters from NVIDIA RTX3090, NVIDIA RTX4090 and NVIDIA RTX5090, respectively.}
\label{tab:throughput}
\end{table}

\textbf{Policy Networks and Input Features.} We evaluate different network architectures and input features on LLaMA-Instruct 3.1 8B, with results detailed in Table~\ref{tab:ablation_inputs}. 
Comparing network architectures, the LSTM-based policy consistently outperforms the MLP-based policy across all datasets, which validates that the draft phase is intrinsically a sequential decision-making process. By capturing temporal dependencies across generation steps, the LSTM architecture achieves more accurate termination decisions than the memoryless MLP.
Regarding input features, using confidence scores yields a higher speedup than using the draft model's hidden states, despite the latter containing richer contextual information. This efficiency stems from a strong positive correlation between confidence scores and token acceptance rates~\cite{li2024eagle2}, which we further confirm by calculating the Pearson correlation of $r=0.4719$ ($p=7.45e^{-137}$) between the top-$1$ score and acceptance length. In contrast, leveraging hidden states demands sufficient semantic understanding, inherently necessitating more training data, extended training time, and a larger prediction model.

\section{Conclusion}

We propose RADAR, a speculative sampling method with RL-based dynamic draft trees. We formulate the draft tree generation as a MDP and employ a prediction model to make real-time decisions of calls to the draft model. To handle the lack of labeled data, which is challenging to obtain due to the randomness and sparsity, we estimate the acceptance length distribution to construct a novel dataset for offline reinforcement learning in training prediction model. Experimental results on three LLMs and four tasks demonstrate that RADAR achieves a speedup of $3.17$x--$4.82$x over vanilla auto-regressive decoding and reduces the calls to the draft model by $18.7\%$ averaged over all tested datasets, outperforming previous methods. 



\section*{Limitations}

While RADAR advances speculative decoding through RL-based dynamic draft trees, it has several limitations that offer avenues for future work. Currently, our method controls only the depth of the draft tree. Extending RADAR to finer-grained topological control, such as per-branch pruning or width adaptation, within an offline RL framework remains an open challenge. Additionally, our prediction model consumes only the top-$k$ confidence scores through a lightweight LSTM. Scaling up the prediction model and incorporating additional input features, such as positional context, may further improve the accuracy of termination decisions, and are left for future work.


\section*{Acknowledgments}
We are greatly appreciative of all the anonymous reviewers for their valuable feedback and insightful suggestions.


\bibliography{custom}

\appendix
\section{RADAR}

\subsection{Inference Pipeline Walkthrough}\label{appendix:pipeline}

Here, we detailed the inference pipeline of our approach by the example shown in Figure~\ref{fig:radar}.
First, consider "\textit{What}" as a query into the LLM, the token "\textit{can}" is generated. Along with some features from the LLM, the token "\textit{can}" is then fed into the draft model for a forward pass, generating two tokens: "\textit{I}" and "\textit{you}", and their confidence scores are returned simultaneously. After that, the prediction model produces a control signal \textit{continue/stop} and a hidden vector $h$ with the confidence scores as inputs. When receiving a \textit{continue} signal, the draft model performs another forward pass with "\textit{I}" and "\textit{you}" as inputs, and produces four tokens: "\textit{say}," "\textit{do}", "\textit{see}" and "\textit{do}", and the top-$k$ confidence scores, corresponding to the tokens "\textit{say}" and "\textit{see}" in Figure \ref{fig:radar} (a). The above steps repeat iteratively. When receiving a \textit{stop} signal, the draft model halts further forward, and the final draft tree is generated, as shown in Figure \ref{fig:radar} (b).

\subsection{Breakdown of the Latency}\label{appendix:latency}
Here we provide a breakdown of the latency in Table \ref{tab:latency}. All data are collected on 2$\times$ NVIDIA RTX3090 GPUs.

\begin{table}[ht]
\centering
\small
\begin{tabular}{lccc}
\toprule
\textbf{Model} & \textbf{$T_o$} & \textbf{$T_d$} & \textbf{$T_{p}$} \\
\midrule
L3 8B & 0.068s & 0.0027s & 0.0004s \\
V 13B & 0.073s & 0.0025s & 0.0004s \\
DSL 8B & 0.0668s & 0.0028s & 0.0004s \\
\bottomrule
\end{tabular}
\caption{Breakdown of the latency for $T_o$, $T_d$, and $T_{p}$. L3 represents LLaMA-Instruct 3.1, V represents Vicuna and DSL represents DeepSeek-R1-Distill-LLaMA.}
\label{tab:latency}
\end{table}

\subsection{Implementation Details}\label{appendix:implementation}
Our prediction model is an LSTM-based network, which consists of a single-layer LSTM followed by a two-layer MLP head. The LSTM outputs are concatenated with the original inputs and projected through the two-layer MLP to produce 2-dimensional logits, which are then transformed into a binary action distribution via Softmax. Detailed parameters of LSTM are used throughout: input\_size=10, hidden\_size=128, num\_layers=1, dropout=0.1. Detailed parameters of MLP are as follows: hidden size=128, ReLU activation, dropout=0.1.

For offline dataset construction, we sample 1,000 prompts from ShareGPT and run EAGLE-3 to collect confidence scores and estimate the acceptance-length distributions. During training, we train the prediction model using REINFORCE with a learning rate of 1e-4, a batch size of 256, for 100 epochs.

\subsection{Interpretation of alpha as a Lagrange Multiplier}
\label{appendix:alpha}

The latency terms $T_{\mathrm{o}}$, $T_{\mathrm{d}}$, and $T_{\mathrm{p}}$ are already included in the terminal throughput reward in Eq.~(\ref{eq:reward}); therefore, $\alpha$ is not an additional latency estimate. Instead, the intermediate penalty supplies a dense learning signal for an otherwise sparse delayed reward, eases credit assignment, and regularizes excessive continuation.

More formally, let $R_T$ denote the terminal throughput reward and let $N_c$ denote the number of \textit{continue} actions in a trajectory. Consider the constrained objective in Eq.~(\ref{eq:constrained_objective}), which maximizes expected terminal throughput subject to an expected draft-computation budget $B$:
\begin{equation}
    \max_{\pi}\; \mathbb{E}_{\pi}[R_T]
    \quad \text{s.t.} \quad
    \mathbb{E}_{\pi}[N_c] \le B.
    \label{eq:constrained_objective}
\end{equation}
Its Lagrangian is given by Eq.~(\ref{eq:lagrangian}):
\begin{equation}
    \mathcal{J}(\pi,\alpha)
    = \mathbb{E}_{\pi}[R_T]
    - \alpha\bigl(\mathbb{E}_{\pi}[N_c]-B\bigr),
    \quad \alpha \ge 0.
    \label{eq:lagrangian}
\end{equation}
For a fixed $B$, optimizing the policy is equivalent, up to the constant $\alpha B$, to maximizing $\mathbb{E}_{\pi}[R_T]-\alpha\mathbb{E}_{\pi}[N_c]$, which is the objective induced by our per-step penalty. Thus, $\alpha$ acts as a Lagrange multiplier that controls the trade-off between terminal throughput and draft computation. Tuning it requires no additional interaction with the target LLM because all candidates reuse the same offline confidence sequences and acceptance-length distributions.

\begin{table*}[ht]
\centering
\small
\begin{tabular}{p{3.4cm}p{5.0cm}p{5.0cm}}
\toprule
\textbf{Aspect} & \textbf{RADAR} & \textbf{LTD} \\
\midrule
RL paradigm & Offline reinforcement learning & Online reinforcement learning \\
Policy model & LSTM-based policy & MLP-based policies \\
Number of policy networks & 1 & 2 \\
Control variables & Draft tree depth via adaptive stopping & Draft tree depth and number of selected draft nodes \\
Training data & Offline dataset generated from 1,000 prompts sampled from ShareGPT & Real-time rollouts from 80 HumanEval samples~\cite{chen2021evaluating} during training \\
LLM interactions during policy optimization & 0 & 100,000 for size policy and 1,000,000 for depth policy \\
Training time & Within 30 minutes & 20--30 GPU hours \\
\bottomrule
\end{tabular}
\caption{Comparison between RADAR and LTD in methodology and training cost.}
\label{tab:radar_ltd_method}
\end{table*}

\subsection{Robustness across Decoding Temperatures}
\label{appendix:temperature}

We further evaluate RADAR with LLaMA-Instruct 3.1 8B under different decoding temperatures. Table~\ref{tab:temperature} reports a controlled temperature sweep. RADAR consistently outperforms EAGLE-3 across all evaluated settings, showing that the learned stopping policy remains effective under both greedy decoding and stochastic sampling.

\begin{table}[ht]
\centering
\small
\begin{tabular}{lccc}
\toprule
\textbf{Dataset} & \textbf{Temperature} & \textbf{EAGLE-3} & \textbf{RADAR} \\
\midrule
GSM8K     & 0.0 & 4.13x & \textbf{4.18x} \\
HumanEval & 0.0 & 4.40x & \textbf{4.48x} \\
MBPP      & 0.0 & 4.71x & \textbf{4.72x} \\
MT-bench  & 0.0 & 4.01x & \textbf{4.12x} \\
MT-bench  & 0.3 & 3.85x & \textbf{3.87x} \\
MT-bench  & 0.5 & 3.74x & \textbf{3.95x} \\
MT-bench  & 0.7 & 3.44x & \textbf{3.51x} \\
MT-bench  & 1.0 & 3.08x & \textbf{3.41x} \\
\bottomrule
\end{tabular}
\caption{Speedup ratios of EAGLE-3 and RADAR under different decoding temperatures with LLaMA-Instruct 3.1 8B.}
\label{tab:temperature}
\end{table}

\subsection{Results on LLaMA-Instruct 3.3 70B}
\label{appendix:70b}

To evaluate scalability on larger models, we further conduct experiments with LLaMA-Instruct 3.3 70B~\cite{dubey2024llama} as the target LLM. These experiments are performed on 2$\times$ RTX PRO 6000 GPUs, and the speedup results are reported in Table~\ref{tab:70b}.
Since the hardware differs from that used in the main experiments, these results serve as a supplementary evaluation.

\begin{table}[ht]
\centering
\small
\begin{tabular}{lcccc}
\toprule
\textbf{Method} & \textbf{MT-bench} & \textbf{GSM8K} & \textbf{Alpaca} & \textbf{MBPP} \\
\midrule
Eagle-3 & 4.42x & 4.83x & 4.93x & 5.35x \\
RADAR & \textbf{4.45x} & \textbf{4.95x} & \textbf{5.07x} & \textbf{5.55x}  \\
\bottomrule
\end{tabular}
\caption{Speedup ratios of Eagle-3 and RADAR across different datasets setting temperature=1 with LLaMA-Instruct 3.3 70B as target LLM on 2$\times$ RTX PRO 6000 GPUs. }
\label{tab:70b}
\end{table}

\subsection{RADAR Effectiveness on HASS}\label{appendix:hass}

To further validate the generality of our approach across different speculative decoding frameworks, we apply RADAR to HASS~\cite{zhang2024learning}, another tree-based speculative decoding method.

\begin{table}[ht]
\centering
\small
\begin{tabular}{lcccc}
\toprule
Method & MT-bench & GSM8K & Alpaca & MBPP \\
\midrule
HASS  & 2.49x & 2.70x & 2.61x & 2.87x \\
RADAR & \textbf{2.56x} & \textbf{2.80x} & \textbf{2.73x} & \textbf{3.03x} \\
\bottomrule
\end{tabular}
\caption{Speedup ratios of HASS and RADAR on LLaMA-Instruct 3.1 8B across four benchmarks on 2$\times$ RTX3090 GPUs setting temperature=1.}
\label{tab:hass_radar}
\end{table}

As shown in Table~\ref{tab:hass_radar}, RADAR consistently improves the speedup of HASS across all four benchmarks on LLaMA-Instruct 3.1 8B~\cite{dubey2024llama}.

\section{Comparison}

\subsection{Comparison with LTD}\label{appdendix:ltd}

We provide a more detailed comparison between RADAR and the concurrent work LTD~\cite{zhang2026learning}, as shown in Table~\ref{tab:radar_ltd_method}. While both methods apply reinforcement learning to adaptive speculative decoding, they differ substantially in policy architecture, training paradigm, data construction, and training cost. In particular, RADAR is built on an offline RL framework, whereas LTD relies on online RL. Consequently, RADAR avoids repeated interaction with the target LLM during policy optimization, since the offline dataset only needs to be constructed once and can thereafter be reused across different devices.

\subsection{Baseline Settings}

For EAGLE-2~\cite{li2024eagle2} and EAGLE-3~\cite{li2025eagle3}, the code and weights are obtained from their official repository\footnote{https://github.com/SafeAILab/EAGLE}. 
For LTD~\cite{zhang2026learning}, the training and evaluation scripts are from its official repository\footnote{https://github.com/zhzihao/Learning-to-Draft}.
For SpecDec++~\cite{huang2024specdec++} and Disco~\cite{mamou2024accelerating}, we train them from scratch using an MLP-based prediction head for 100 epochs. Both models are trained with the AdamW optimizer~(lr=1e-4).

\section{Licenses of Artifacts}
The artifacts used in this work are licensed as follows: LLaMA-Instruct 3.1 8B/3.3 70B (LLAMA 3.1 COMMUNITY LICENSE/LLAMA 3.3 COMMUNITY LICENSE), Vicuna 13B v1.3 (Non-commercial license), DeepSeek-R1-Distill-LLaMA (MIT License), EAGLE-2/3 and HASS (Apache 2.0), MT-bench (Apache 2.0), GSM8K (MIT License), Alpaca (CC BY-NC 4.0), MBPP (CC BY 4.0), and ShareGPT (Apache 2.0). All artifacts are used strictly for non-commercial academic research, consistent with their respective licenses. RADAR is also intended solely for academic research on speculative decoding and LLM inference acceleration.

\end{document}